\documentclass[pdflatex,sn-mathphys-num]{sn-jnl}

\usepackage{graphicx}%
\usepackage{multirow}%
\usepackage{amsmath,amssymb,amsfonts}%
\usepackage{amsthm}%
\usepackage{mathrsfs}%
\usepackage[title]{appendix}%
\usepackage{xcolor}%
\usepackage{textcomp}%
\usepackage{manyfoot}%
\usepackage{booktabs}%
\usepackage{algorithm}%
\usepackage{algorithmicx}%
\usepackage{algpseudocode}%
\usepackage{listings}%
\theoremstyle{thmstyleone}%
\theoremstyle{thmstyletwo}%

\theoremstyle{thmstylethree}%

\colorlet{mutedForegroundColor}{gray!80!black}
\colorlet{mutedBackgroundColor}{gray!10}
\newcommand{\mutedColor}{gray}

\colorlet{primaryForegroundColor}{blue!80!black}
\colorlet{primaryBackgroundColor}{blue!10}

\definecolor{secondaryForegroundColor}{RGB}{34, 139, 34}
\colorlet{secondaryBackgroundColor}{secondaryForegroundColor!10}

\colorlet{tertiaryForegroundColor}{red}
\colorlet{tertiaryBackgroundColor}{red!10}
\newcommand{\tertiaryColor}{red}

\colorlet{quaternaryForegroundColor}{orange}
\colorlet{quaternaryBackgroundColor}{orange!10}

\newcommand{\CC}{C\nolinebreak\hspace{-.05em}\raisebox{.4ex}{\tiny\bf +}\nolinebreak\hspace{-.10em}\raisebox{.4ex}{\tiny\bf +}}
\def\CC{{C\nolinebreak[4]\hspace{-.05em}\raisebox{.4ex}{\tiny\bf ++}}}

\begin{document}
    \title[Article Title]{SFG-ROS: A Resource-Aware Framework for Dense Multi-Agent Perception}
    
    
    \author*{\fnm{Constantin} \sur{Blessing}}\email{constantin.blessing@hs-esslingen.de}
    \author{\fnm{Elias} \sur{Geiger}}\email{elias.geiger@hs-esslingen.de}
    \author{\fnm{Jakob} \sur{Häringer}}\email{jakob.haeringer@hs-esslingen.de}
    \author{\fnm{Dennis} \sur{Grewe}}\email{dennis.grewe@hs-esslingen.de}
    \author{\fnm{Markus} \sur{Enzweiler}}\email{markus.enzweiler@hs-esslingen.de}

   \affil{\orgdiv{Institute for Intelligent Systems, Faculty of Computer Sciences and Engineering}, \orgname{Esslingen University of Applied Sciences}, \orgaddress{\city{Esslingen am Neckar}, \postcode{73732}, \state{Baden-Württemberg}, \country{Germany}}}

    
    \abstract{Deploying heterogeneous multi-agent robot fleets for collaborative perception requires robust data exchange and scalable software architectures. However, standard ROS 2 implementations often suffer from network saturation, namespace collisions, and severe computational overhead when distributing dense sensor streams across devices. To address these bottlenecks, we present SFG-ROS, a resource-aware multi-agent software framework designed for dynamic fleet deployments. SFG-ROS addresses these challenges through three primary contributions. First, schema-driven traffic routing isolates high-frequency intra-agent traffic from the global network using a programmatic fully qualified name schema and targeted Fast DDS routing. Second, an on-demand centralized decoding pipeline automatically offloads high-bandwidth sensor data decompression, eliminating redundant processing across local consumer nodes. Finally, a hardware-agnostic container pipeline dynamically adapts to heterogeneous accelerators, seamlessly bridging development environments with zero-touch, field-ready execution. We evaluate the framework using a fleet of wheeled and legged robots equipped with LiDAR and stereo depth cameras. Experimental results show SFG-ROS bounds network traffic to $\mathcal{O}(1)$ and, by replacing redundant decompression with lightweight IPC, reduces the per-subscriber CPU scaling penalty by 72.3\% versus standard ROS 2, all while maintaining low latency. Finally, we publish SFG-ROS under a permissive license, available via \href{https://iis-esslingen.github.io/sfg-ros}{iis-esslingen.github.io/sfg-ros}.}

    
    
    
    \maketitle

    \section{Introduction} \label{section:introduction}
    
    Industry 4.0 has driven the development of dynamic manufacturing facilities. Traditional mass production, characterized by rigid assembly lines and static material flow, is increasingly being augmented or replaced by highly flexible concepts such as matrix production \cite{MOHR20241972}. Pushing this paradigm further, the vision of Smart Factory Grids (SFG) \cite{sfgWebsite} proposes dynamically distributed production networks composed of specialized, autonomous cyber-physical systems. The SFG vision aims to enable highly resilient, flexible manufacturing for low-quantity rapid prototyping with minimal setup times.

    In this context, heterogeneous mobile robots -- spanning wheeled, legged, aerial, and more recently humanoid platforms \cite{malikIntelligentHumanoid} -- serve as the connective tissue of the SFG, enabling material flow, automated facility inspection, and cooperative maintenance necessary for distributed production. To function effectively, these heterogeneous robots (constituting the multi-agent system) must operate not as isolated entities, but as a collaborative fleet capable of sharing high-bandwidth sensor data, dynamic state information, and task-level intentions in real time. While these smart factory deployments serve as our primary motivating use case, the fundamental challenge of coordinating such fleets is universal to any multi-agent robotic system.

    Accordingly, facilitating collaborative tasks requires a robust software foundation that governs hardware abstraction, task execution, and inter-process communication. While the Robot Operating System 2 (ROS 2) \cite{ros2} is the standard robotics framework, its default configuration limits deployments on resource-constrained robot compute units (edge devices) in dense, multi-agent networks. Specifically, decentralized discovery requires a fully-connected graph, causing network-saturating packet storms as fleets scale \cite{roscoreteam2023rmw}. Furthermore, sharing dense sensor streams like images overwhelms default UDP buffers \cite{roscoreteam2023rmw}; mitigating this via \verb|image_transport| compression merely shifts the bottleneck, exhausting CPU resources through redundant decompression across local subscribers. Finally, heterogeneous hardware (e.g., ARM, CUDA, ROCm) creates significant deployment friction. To address these bottlenecks, we derive four core requirements for an edge-assisted, multi-agent framework against which existing solutions are evaluated:
    
    \begin{enumerate}
        \item \label{requirement:trafficAndNamespaceScalability} \textbf{Traffic and Namespace Scalability:} Prevent network saturation and namespace collisions by ensuring high-frequency local operations do not congest the global network, while preserving reliable inter-agent communication.
        \item \label{requirement:resourceEfficientPerception} \textbf{Resource-Efficient Perception:} Minimize compute and bandwidth overhead by eliminating redundant sensor processing on devices and avoiding continuous multicast discovery storms.
        \item \label{requirement:platformAgnosticPortability} \textbf{Platform-Agnostic Portability:} Ensure reproducible, zero-touch deployment across heterogeneous hardware architectures to seamlessly bridge development and production environments.
        \item \label{requirement:fleetWideObservability} \textbf{Fleet-Wide Observability:} Enable unified multi-agent coordination by seamlessly aggregating distributed state data, synchronizing spatial reference frames, and providing coherent temporal visualization.
    \end{enumerate}
    
    \section{Related Work}

    This section reviews prior work across three key areas: multi-agent ROS 2 frameworks, edge/cloud robotics offloading, and deployment and virtualization.
    
    \subsection{Multi-Agent ROS 2 Frameworks}

    Several frameworks address the complexities of coordinating multiple ROS 2 agents. Open-RMF \cite{openRmf} is widely utilized to facilitate interoperability and task-level traffic management across heterogeneous fleets. However, it primarily operates at the orchestration layer to achieve unified coordination (Requirement \ref{requirement:fleetWideObservability}). It does not inherently resolve the low-level middleware bottlenecks associated with decentralized discovery storms and namespace collisions, falling short of the strict traffic isolation needed for network scalability (Requirement \ref{requirement:trafficAndNamespaceScalability}). Similarly, RoboFleet \cite{sikand2021RoboFleet} addresses communication overhead by bridging distributed ROS networks over WebSockets. While effective for wide-area monitoring, it requires protocol translation and introduces a central point of failure by utilizing a single-server architecture for dense, high-frequency multi-agent perception sharing.
    
    \subsection{Edge/Cloud Robotics Offloading}

    The computational burden of dense multi-agent perception has driven the development of various offloading frameworks. FogROS 2 \cite{ichnowski2023FogRos2}, for instance, extends the ROS 2 ecosystem by enabling the seamless migration of compute-heavy nodes to external edge or cloud servers. While cloud offloading circumvents edge-compute limitations, it inherently relies on persistent, high-bandwidth external links. Furthermore, the geographical sparsity of cloud servers can induce latency boundaries unsuitable for collaborative perception systems that require continuous, data-intensive sensor streams. Consequently, it fails to provide the localized, resource-efficient perception (Requirement \ref{requirement:resourceEfficientPerception}) necessary to eliminate redundant onboard processing without offloading data across volatile networks.

    \subsection{Deployment \& Virtualization}
    
    Transitioning ROS 2 applications from development to production across diverse robotic hardware necessitates robust virtualization. Given the heavy reliance of ROS 2 on specific OS-level dependencies and hardware drivers, lightweight containerization (e.g. Docker) has emerged as the practical standard \cite{ros2_deployment_guidelines,ros2,pr14050804}. Solutions leveraging Kubernetes, such as KubeROS \cite{zhang2023kuberos}, bring cloud-native orchestration to robotic fleets but often introduce management and resource overhead per agent (e.g. a kubelet node agent) that is prohibitive for severely constrained edge platforms. Conversely, vendor-specific solutions like NVIDIA Isaac ROS \cite{nvidiaIsaacRos} provide highly optimized, containerized deployments with deep hardware acceleration. However, these solutions inherently lock deployments into a single hardware ecosystem (e.g. CUDA), failing to meet the platform-agnostic portability (Requirement \ref{requirement:platformAgnosticPortability}) necessary for zero-touch deployment across a heterogeneous fleet containing various devices and competing hardware accelerators.
    
    \section{Contributions} \label{section:contributions}
    
    We introduce SFG-ROS, a novel ROS 2-based multi-agent software framework designed to fulfill the core requirements established in Section \ref{section:introduction} for resource-constrained collaborative perception and fleet coordination. Our primary contributions are:

    \begin{enumerate}
        \item \textbf{Schema-Driven Traffic Routing:} A topology-enforcing communication architecture that utilizes a programmatic fully qualified name schema and targeted communication routing to isolate intra-agent from inter-agent traffic, fulfilling the need for traffic and namespace scalability (Requirement \ref{requirement:trafficAndNamespaceScalability}).

        \item \textbf{On-Demand Stream Decompression:} A dynamic agent discovery and centralized data decoding pipeline that significantly reduces network and computational overhead by selectively decompressing sensor streams on demand, enabling resource-efficient perception (Requirement \ref{requirement:resourceEfficientPerception}).
        
        \item \textbf{Accelerator-Aware Virtualization:} A hardware-agnostic container deployment strategy that dynamically adapts to heterogeneous accelerators for a seamless transition from development to deployment, ensuring platform-agnostic portability (Requirement \ref{requirement:platformAgnosticPortability}).
        
        \item \textbf{Spatiotemporal Synchronization \& Visualization:} A suite of multi-agent visualization and planning tools, including a visualizer bridge and a proxy-frame external ground truth synchronizer, which together provide fleet-wide observability (Requirement \ref{requirement:fleetWideObservability}).
    \end{enumerate}

    The remainder of this paper is structured as follows: Section \ref{section:architectureAndImplementtaion} details the proposed SFG-ROS architecture and implementation. Section \ref{section:experimentalEvaluation} presents the experimental evaluation. Finally, Section \ref{section:ConclusionAndFutureWork} concludes the paper and outlines directions for future work.

    \section{Architecture \& Implementation} \label{section:architectureAndImplementtaion}
    
    \begin{figure}[htbp]
        \centering
        \includegraphics[width=\textwidth]{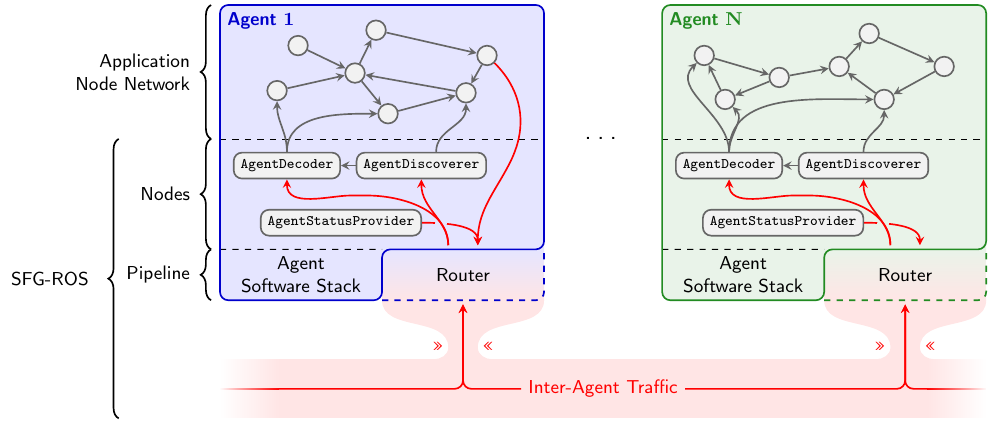}
        \caption{The SFG-ROS architecture comprised of the underlying pipeline and dedicated inter-agent communication nodes. A dedicated router bridges raw intra-agent (\mutedColor{} arrows) and compressed inter-agent (\tertiaryColor{} arrows) traffic based on our topologically enforced communication scheme}\label{figure:architecture}%
    \end{figure}

    The proposed framework aims to transform conceptual multi-agent coordination into a high-performance, reproducible software stack. Our primary objective is to mitigate the inherent overhead of ROS 2 discovery and data transport in dense sensor environments without sacrificing the middleware's modularity and flexibility. To achieve this, we introduce a ``global-compressed, local-raw" architectural paradigm, as illustrated in Figure \ref{figure:architecture}. Under this paradigm, high-bandwidth data is strictly encapsulated and compressed for inter-agent transport across a global bridge domain, while internal device processing within the local domain remains raw, transparent, and computationally efficient.

    In the following subsections the implementation of this architecture in alignment with the core contributions from Section \ref{section:contributions}, progressing from the foundational routing schema and data exchange pipelines to deployment strategies and fleet visualization tools, will be detailed.
    
    \subsection{Communication Topology \& Naming Convention} \label{subsection:communicationTopologyAndNamingConvention}
    
    By default, ROS 2 agents rely on the Data Distribution Service (DDS), a decentralized middleware specification, to automatically discover one another on a local network and to exchange data through topics, services, and actions \cite{ros2}. To minimize network interference and namespace collisions in our multi-agent setup -- in accordance with Requirement \ref{requirement:trafficAndNamespaceScalability} -- we employ Fast DDS by eProsima \cite{eprosimaFastDds} as our middleware implementation alongside a two-layered communication isolation approach.
    
    First, each agent is assigned a unique ROS domain ID to ensure that DDS network traffic is segregated between agents on the DDS layer itself. While this is an effective measure to prevent accidental cross-agent talk, it prohibits all communication between agents, even desired one necessary for collaborative tasks. Therefore, to facilitate cross-agent communication, each agent makes use of a middleware routing mechanism (implemented here via a Fast DDS Router \cite{eprosimaDdsRouter}) that selectively routes middleware traffic between the agent's domain and a shared, global domain based on the communication's fully qualified name (FQN).

    \begin{table}[htbp]
        \centering
        \caption{DDS router Forwarding Rules. Note that the FQN prefixes for the forwarding rules follow the ROS 2 DDS mapping convention (e.g. \texttt{rt/} for topics)}\label{table:ddsRouterRules}%
        \begin{tabular}{lll}
            \toprule
            \textbf{Forwarding Rule} & \textbf{Participating Domains} & \textbf{Description} \\
            \midrule
            \verb|rt/tf| & \multirow{5}{*}{\begin{tabular}{@{}c@{}} $\vdots$ \\ $D_i$/$D_0$ \\ $\vdots$ \end{tabular}} & Transform topic \\
            \verb|rt/tf_static| & & Static transform topic \\
            \verb|rt/global/*|  & & Global topics \\
            \verb|rq/global/*|  & & Global service requests \\
            \verb|rr/global/*|  & & Global service responses \\
            \bottomrule
        \end{tabular}
    \end{table}
    
    Formally, let $D_i \neq 0$ denote the agent's local domain for a specific agent $i$, $D_0 = 0$ the global domain, and $\mathcal{F}_{\text{forward}}$ the set of all possible FQNs that satisfy the forwarding rules outlined in Table \ref{table:ddsRouterRules}. The routing policy $\pi$ for a given message $m$ published on a topic with FQN $f$ is enforced as a piecewise mapping
    \begin{equation}
        \pi(m, f) = 
        \begin{cases} 
        D_i \leftrightarrow D_0 & \text{if} f \in \mathcal{F}_{\text{forward}} \\
        D_i & \text{otherwise}
        \end{cases}
    \end{equation}

    As seen in Table \ref{table:ddsRouterRules} and defined by the policy $\pi$, a given DDS router forwards messages bidirectionally between the dedicated agent's domain $D_i$ and the domain $D_0$, transforming the latter into a global bridge domain that acts as the hub for cross-agent communication. Note that for transform topics, appropriate frame prefixing is required to prevent TF frame ID clashes.

    Second, to guarantee that developers comply with these routing rules, we implement a purpose-built, code-enforced FQN naming convention. We define a valid FQN as an ordered tuple $T = (s, a, c, d, r)$, constructed from individual segments:
    \begin{itemize}
        \item $s \in \mathcal{S}$ is the mandatory scope segment.
        \item $a \in \mathcal{A} \cup \{ \emptyset \}$ is the optional agent segment.
        \item $c \in \mathcal{C} \cup \{ \emptyset \}$ is the optional component segment.
        \item $d \in \mathcal{D} \cup \{ \emptyset \}$ is the optional data stream segment.
        \item $r \in \mathcal{R}$ is the mandatory resource segment.
    \end{itemize}

    To bridge these formal sets with the software implementation, the instantiation of $T$ is managed by a \verb|RosFqnBuilder| class utilizing the builder pattern. Rather than relying on error-prone manual string concatenation, the builder is supplied with programmatic representations of the tuple segments. The sets $\mathcal{S}$, $\mathcal{C}$, $\mathcal{D}$, and $\mathcal{R}$ are defined in the codebase via corresponding enumerations containing standardized values, such as \verb|Lidar| or \verb|LocomotionController| in the case of set $\mathcal{C}$. In contrast, the agent set $\mathcal{A}$ is resolved dynamically at runtime: the builder defaults to a sanitized version of the operating system's hostname, which can be explicitly overridden by setting the \verb|SFG_AGENT_NAME| environment variable. 

    \begin{figure}[htbp]
        \centering
        \includegraphics[width=\textwidth]{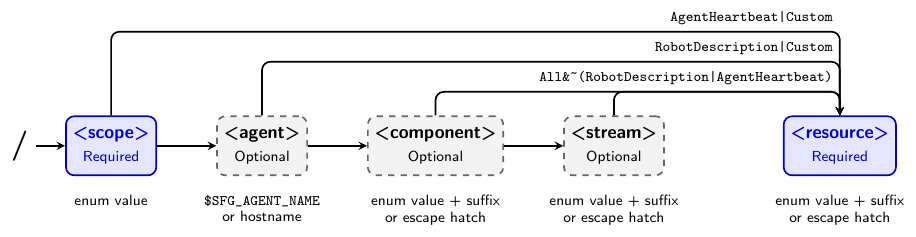}
        \caption{Structural syntax of the programmatic FQN schema. The paths leading into the mandatory \texttt{<resource>} segment are annotated with their bitwise constraints, dictating which resource types are permitted based on the preceding segment of the FQN}\label{figure:fqnSchema}%
    \end{figure}
    
    To maintain flexibility, the implementation provides two extension mechanisms for the enumeration-backed sets $\mathcal{C}$, $\mathcal{D}$, and $\mathcal{R}$. First, an optional suffix can be appended to a given enumeration value to provide additional granularity (e.g. differentiating between a front and a rear camera). Second, an escape hatch is available via a dedicated \verb|Custom| enumeration value. This allows developers to assign an arbitrary string to a segment to accommodate novel use cases, provided the resulting string remains a valid ROS FQN. The set $\mathcal{S}$ strictly encompasses the two distinct values \verb|Local| and \verb|Global|.
    
    Ultimately, the construction of the valid ROS 2 FQN from $T$ is achieved by the builder automatically concatenating the resolved tuple elements using standard ROS namespace separators. The structural syntax and dependency paths of this schema are illustrated in Figure \ref{figure:fqnSchema}.

    \subsection{Dynamic Discovery \& Agent Lifecycle} \label{subsection:dynamicDiscoveryAndAgentLifecycle}

    Ensuring traffic and namespace scalability (Requirement \ref{requirement:trafficAndNamespaceScalability}) necessitates going beyond the ROS 2-native discovery mechanisms (multicast UDP). Therefore, we deploy a standardized set of lifecycle management nodes that serve as the fundamental baseline for every agent in the multi-agent system.

    The \verb|AgentStatusProvider| node broadcasts \verb|AgentHeartbeat| messages on the globally routed topic \verb|"/global/agent_heartbeat"| at periodic intervals, defaulting to 1s during agent startup and 5s during steady state operation in our implementation. Simultaneously, it provides a global service \verb|"/global/<agent>/get_metadata"| that allows other agents to query its specific hardware configuration, such as available sensors.

    \begin{figure}[htbp]
        \centering
        \includegraphics{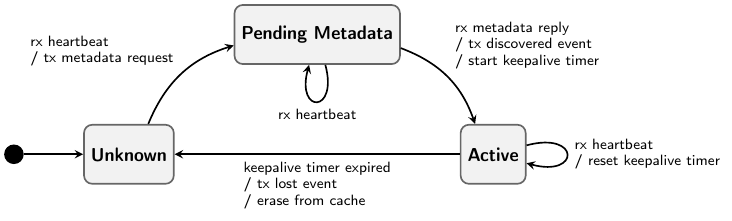}
        \caption{The internal lifecycle state machine of the \texttt{AgentDiscoverer} node for a given remote agent, using standard UML syntax (\texttt{Event/Action}). The node leverages asynchronous ROS 2 service futures and timers to track agents and prevent redundant metadata queries, exposing the resulting state to the local domain via discovery events}\label{figure:agentLifecycleStateMachine}%
    \end{figure}
    
    The \verb|AgentDiscoverer| node acts as a centralized abstraction layer on top of the raw heartbeats. By subscribing to the global heartbeat, it maintains an active cache of the fleet and fires agent discovery events via the locally scoped topic \verb|"/local/<agent>/agent_discovery_event"|. This approach strictly bounds complex discovery logic and asynchronous service calls to a single node per agent, shielding local consumer nodes from network-level state management. The internal lifecycle state machine for tracking a remote agent, modeled as a Deterministic Finite Automaton, is illustrated in Figure \ref{figure:agentLifecycleStateMachine}.
    
    \subsection{Efficient Sensor Data Exchange} \label{subsection:efficientSensorDataExchange}

     For Requirement \ref{requirement:resourceEfficientPerception}, i.e., efficiently sharing dense, high-frequency camera data, such as depth or color images, compression is required. To ensure compatibility with third-party camera drivers, we leverage the standard ROS 2 \verb|image_transport| interface. For color imagery, we utilize the FFmpeg plugin by \cite{ffmpegImageTransport} to enable hardware-accelerated H.264 encoding. FFmpeg's broad selection of image codecs enables the utilization of platform-specific hardware accelerators (e.g. NVENC/NVDEC) without requiring modifications to application-level code. 
     
     Because standard video codecs introduce artifacting that corrupts radiometric depth data \cite{merkle2009,tech2009}, and the available, lossless PNG compression is comparatively slow, we backported the Fast Lossless Depth Video, RVL compression algorithm \cite{rvlCompression} from later ROS 2 distributions to our Humble-based stack. Furthermore, all sensor streams enforce best effort quality of service (QoS) reliability profiles to prevent stale data buildup and retransmission induced latency spikes.
     
    A significant bottleneck inherent to the standard \verb|image_transport| architecture arises when multiple consumer nodes residing on the same device simultaneously subscribe to a single remote compressed image stream. By default, this triggers independent, redundant decompression steps for every subscriber, leading to severe computational overhead. To mitigate this, we implemented an \verb|AgentDecoder| node that acts as a centralized, dynamic proxy for decoding, reducing the decoding overhead per physical device.
    
    Let $\mathcal{S}$ be the set of subscribers to a specific compressed sensor stream $r$. In a base ROS 2 architecture, the total computational cost of decompression $c_{\text{base}}(r)$ on a single device scales linearly with the number of subscribers, i.e. 
    \begin{equation}
        c_{\text{base}}(r) = \sum_{i=1}^{|\mathcal{S}|} \big( c_{\text{net}}(r) + c_{\text{dec}}(r) \big),
    \end{equation}
    where $c_{\text{net}}(r)$ is the network protocol overhead and $c_{\text{dec}}(r)$ is the CPU cost of e.g. decoding H.264 or RVL frames. By introducing the \verb|AgentDecoder|, the decompression operation is decoupled from the consumers. The cost model simplifies to
    \begin{equation}
        c_{\text{proposed}}(r) = c_{\text{net}}(r) + c_{\text{dec}}(r) + \sum_{i=1}^{|\mathcal{S}|} c_{\text{ipc}}(r),
    \end{equation}
    where $c_{\text{ipc}} \ll c_{\text{dec}}$ represents the cost of inter-process communication (IPC). This decoupling reduces the per-subscriber computational penalty to strictly IPC overhead, resulting in a flatter linear scaling curve for CPU utilization while maintaining $\mathcal{O}(1)$ network bandwidth.

    \begin{figure}[htbp]
        \centering
        \includegraphics[width=\textwidth]{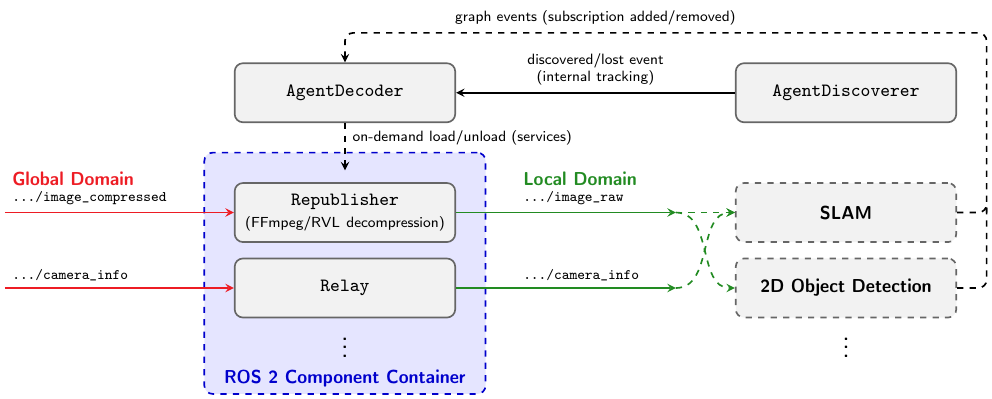}
        \caption{Architecture of the dynamic, centralized decoding pipeline. The \texttt{AgentDecoder} registers available remote agents via discovery events but relies on ROS 2 graph events to monitor local domain demand}\label{figure:agentDecoder}%
    \end{figure}
    
    As illustrated in Figure \ref{figure:agentDecoder}, the \verb|AgentDecoder| subscribes to local agent discovery events. Upon discovering a new agent, it registers the agent's metadata but defers the instantiation of any decompression pipelines. Instead, the \verb|AgentDecoder| actively monitors ROS 2 graph events to track local subscriptions. When a local consumer node subscribes to a specific decoded or relayed topic associated with a known active agent, the \verb|AgentDecoder| leverages ROS 2 \verb|composition_interfaces| to dynamically load the required components directly into a shared ROS 2 component container. Consequently, when all local subscribers for a specific topic disconnect, the \verb|AgentDecoder| automatically unloads the corresponding component, ensuring deterministic, demand-driven resource reclamation and preventing wasted compute on unutilized data streams.
    
    This architecture manages the on-demand decompression of images and the relaying of \verb|CameraInfo| topics respectively, and can be extended to other compressed sensor streams. By utilizing composable nodes within a unified container process, we minimize operating system context-switching and lay the groundwork for zero-copy intra-process communication available in future ROS 2 distributions.
    
    \subsection{Deployment \& Development Pipeline}

    \begin{figure}[htbp]
        \centering
        \includegraphics[width=\textwidth]{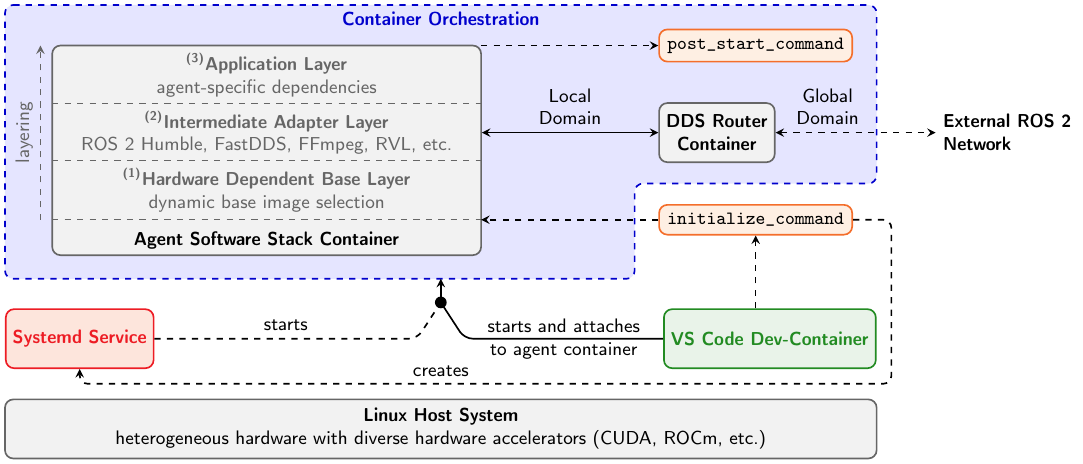}
        \caption{Overview of the dual-container deployment and development architecture. A modular, three-layer agent container adapts to heterogeneous Linux host hardware, while a dedicated DDS router container bridges isolated agent communications with the external ROS 2 network. The stack can be started either via VS Code or a systemd service for automatic startup} \label{figure:deploymentAndDevelopmentPipeline}%
    \end{figure}
    
    To ensure reproducibility across heterogeneous hardware -- ranging from NVIDIA Jetson-based robots to ARM-based PC Handhelds -- we implement a layered, containerized deployment strategy depicted in Figure \ref{figure:deploymentAndDevelopmentPipeline}.

    The software stack is built upon a modular, multi-stage Dockerfile architecture. We employ a dynamic base image selection process that detects the host's hardware accelerators (e.g. CUDA for NVIDIA, ROCm for AMD, etc.) during the build process. The architecture is composed of three distinct layers, namely (1) the hardware-dependent base layer, (2) the intermediate adapter layer establishing a common functional baseline including ROS 2 Humble, FFmpeg, RVL decompression, etc., and (3) the application layer encompassing agent-level dependencies and the agent's software stack via \verb|colcon| workspace mounting.
    
    Deployment on the physical agent is managed via a dual-container orchestration using Docker Compose. The first container encapsulates the agent software stack using the aforementioned architecture, while the second runs the dedicated DDS router instance mentioned in Section \ref{subsection:communicationTopologyAndNamingConvention}. For deployment, we provide systemd service integration, enabling the entire stack to initialize automatically upon agent power-on. Lastly, the environment is integrated with VS Code through specialized dev-container configurations, providing:

    \begin{itemize}
        \item \textbf{Tooling:} Preconfigured IntelliSense for the colcon workspace, alongside standardized formatting for \CC{}, Python, and CMake.
        
        \item \textbf{Debugging:} Integrated debugging profiles for both \CC{} and Python nodes with full CLI argument forwarding.
        
        \item \textbf{Lifecycle Hooks:} Agent-specific commands such as the \verb|initialize_command| executed during dev-container setup which handles persistent host-to-container configurations, such as establishing NTP-based time synchronization, configuring firewalls, and generating the aforementioned systemd service. To maintain flexibility, all hooks support custom user extensions via dedicated extension scripts.
    \end{itemize}

    \subsection{Specialized Tooling}

    To complement the core framework and deployment architecture, we have developed a suite of specialized tools designed for visualization and coordination (Requirement \ref{requirement:fleetWideObservability}).

    The standard ROS 2 visualization tools such as RViz \cite{kam2015RViz} often struggle with the spatiotemporal complexity and historical logging of dense, multi-agent networks. Conversely, high-performance visualization frameworks like the Rerun data viewer \cite{RerunSDK} do not feature native ROS 2 integration. 
    
    Addressing this, we implement a bridge for the Rerun data viewer, designed with an RViz-like graphical user interface to dynamically add visualizers, configure QoS profiles, and tune visualizer-specific parameters at runtime. The architecture relies on \verb|pluginlib| to decouple the core bridge logic from message-specific handling, supporting the creation of custom visualizer plugins for both standard and non-standard ROS 2 messages. Notably, the plugin interface natively supports ROS 2 message filtering and synchronization. This allows developers to easily create complex visualizers that fuse multiple dependent topics, such as precisely synchronizing \verb|Image| streams with their respective \verb|CameraInfo| for accurate 3D depth map projection, requiring solely the implementation of a unified data-mapping callback.

    \begin{figure}[htbp]
        \centering
        \includegraphics[width=\textwidth]{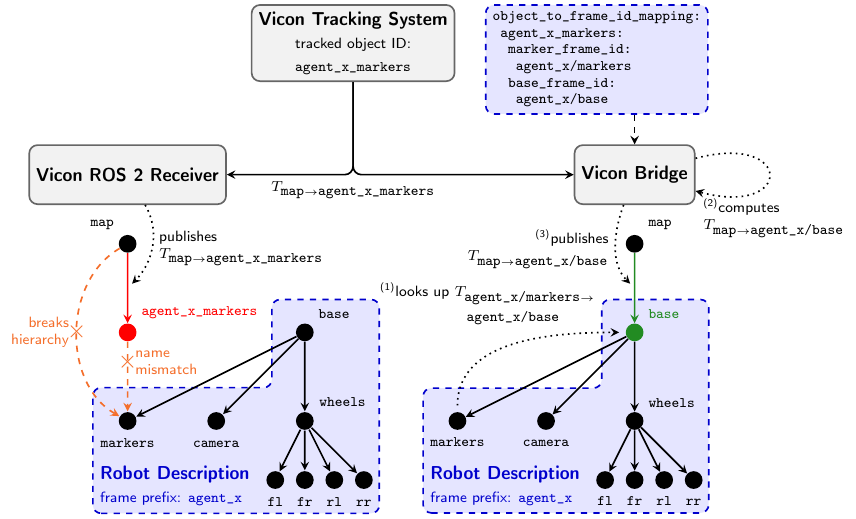}
        \caption{Comparison of TF tree integration for Vicon motion capture data, contrasting the official Vicon ROS 2 Receiver (left) with our Vicon Bridge (right)}\label{figure:viconBridge}%
    \end{figure}

    While the Rerun bridge provides temporal, visual logging, experimental evaluation as well as perception itself requires precise pose information. In our specific use-case we leverage a Vicon Vero 2.2 \cite{viconVero} camera system using the Vicon Tracker 4.4 \cite{viconTracker} software for external agent localization via reflective markers placed on the agents themselves. While Vicon officially endorses the Vicon ROS 2 Receiver \cite{viconReceiverForRos2}, it proved unsuitable for our application due to two limitations: (1) It rigidly maps Vicon tracked object IDs to ROS 2 TF frame IDs with no remapping mechanism in place, and (2) it is incompatible with robot descriptions whose root does not equal the tracked object itself, resulting in violation of the single root requirement for ROS 2 TF trees. Therefore, we propose Vicon Bridge that solves both issues by (1) allowing for an arbitrary mapping between Vicon object IDs and ROS 2 TF frame IDs, and (2) employing a frame-via-proxy-frame-publication mechanism. 
    
    Figure \ref{figure:viconBridge} contrasts both approaches: As a direct consequence of the first limitation, the official receiver produces a name mismatch between the published frame \verb|agent_x_markers| and the target frame \verb|agent_1/markers|. Furthermore, even if this naming conflict were resolved, the aforementioned root discrepancy inherently breaks the TF hierarchy, since \verb|agent_x|'s accompanying robot description is rooted at \verb|base|, not at \verb|markers|.
    
    In contrast, our Vicon Bridge's frame-via-proxy-frame-publication mechanism publishes the desired transform $T_{\texttt{map} \to \texttt{agent\_x/base}}$, resolving both issues. It achieves this by (1) looking up the transform $T_{\texttt{agent\_x/markers} \to \texttt{agent\_x/base}}$ followed by (2) computing the transform
    \begin{equation}
        T_{\texttt{map} \to \texttt{agent\_x/base}} = T_{\texttt{map} \to \texttt{agent\_x\_markers}} \cdot T_{\texttt{agent\_x/markers} \to \texttt{agent\_x/base}}
    \end{equation}
    and (3) concluding with the subsequent publication of the result.

    \begin{figure}[htbp]
        \centering
        \includegraphics[width=\linewidth]{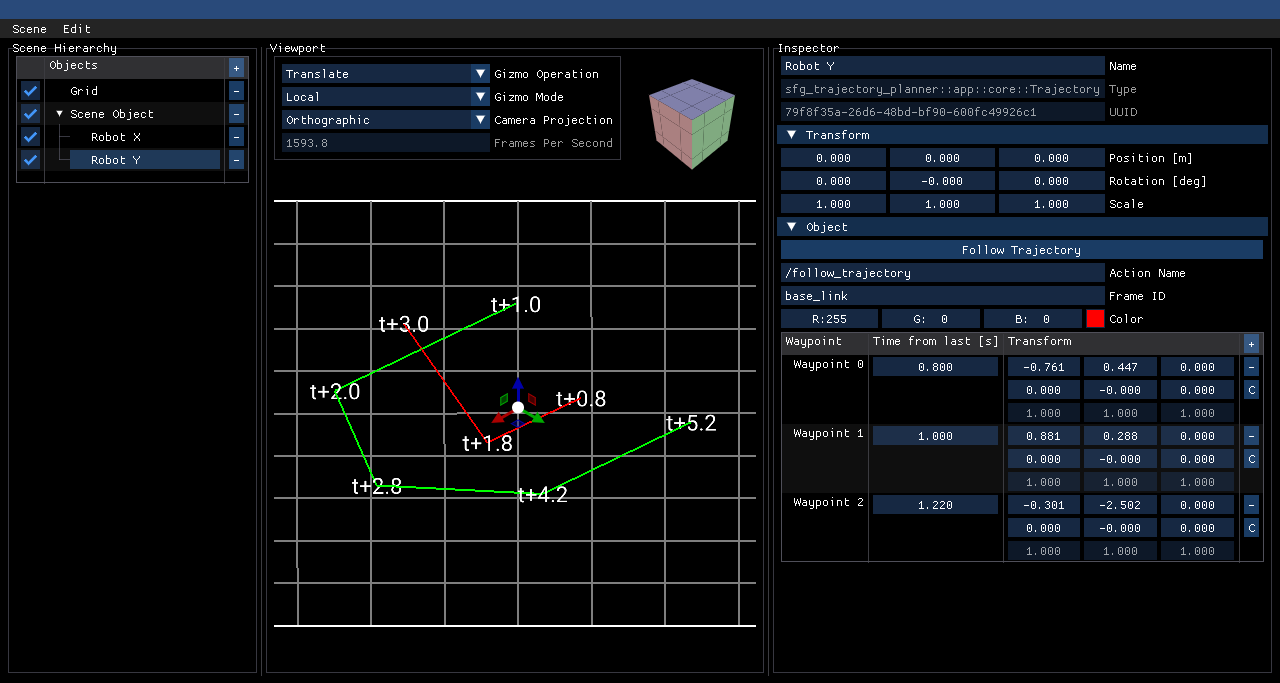}
        \caption{User interface of the multi-agent trajectory planner, featuring a scene graph hierarchy (left), an interactive 3D viewport (center), and a property inspector (right)}\label{figure:trajectoryPlanner}%
    \end{figure}
    
    Lastly, to translate synchronized perception and localization into collaborative action, the ecosystem features a multi-agent trajectory planner for $\text{SE}(3)$ pose-based navigation. The planner operates on a scene graph, encompassing scene objects, and an associated editor featuring a scene hierarchy, a 3D viewport as well as a scene object inspector depicted in Figure \ref{figure:trajectoryPlanner}. The latter also includes undo/redo functionality as well as YAML-based scene serialization. Mathematically, the planner models a trajectory $\mathcal{T}$ as an ordered set of $N$ waypoints $\mathcal{T} = \{ (\mathbf{p}_i, t_i) \}_{i=1}^N$ where each waypoint consists of an $\text{SE}(3)$ spatial pose $\mathbf{p}_i$ and a corresponding temporal offset $t_i \in \mathbb{R}_{\geq 0}$. To streamline waypoint orientation, the planner offers automated alignment heuristics, allowing poses to automatically align with the previous or next waypoint. While the current iteration relies on piecewise polyline interpolation, future development will integrate centripetal Catmull-Rom splines \cite{barry1988Recursive} to generate smooth, kinematically feasible paths. For execution, the trajectory inspector allows the direct publishing of trajectories to a configurable ROS 2 action server.

    \section{Experimental Evaluation} \label{section:experimentalEvaluation}

    We evaluate the efficiency and scalability of SFG-ROS by benchmarking our architecture against a standard ROS 2 decentralized deployment. We define and focus on three key evaluation metrics:
    
    \begin{itemize}
        \item \textbf{Bandwidth \& Resource Optimization:} We measure CPU utilization and network bandwidth on a camera-data-consuming agent as the number of local subscribers increases, to assess the \verb|AgentDecoder|'s efficacy.
        \item \textbf{System Dynamics \& Discovery:} We evaluate network scalability of the fleet by measuring discovery time and network traffic as the number of agents scales, to quantify the benefits of isolating inter- and intra-agent traffic.
        \item \textbf{Latency Overhead:} We measure glass-to-glass latency from initial camera capture by an edge agent to visual rendering on a remote consumer agent and the subscription-to-first-frame latency incurred by on-demand resource allocation, to assess the delay introduced by \verb|AgentDecoder| instantiation.
    \end{itemize}
    
    \subsection{Experimental Setup} \label{subsection:experimentalSetup}

    The multi-agent testbed consists of a heterogeneous deployment bridging edge robotics, a fleet of single-board computers, a mobile teleoperation device, and a high-performance base station. The primary edge agent is a Unitree Go2 legged robot equipped with an NVIDIA Jetson Orin NX (16GB) compute unit and a USB-C connected Intel RealSense D435 camera. For bandwidth and resource optimization, we utilize a workstation as a base station featuring an AMD Ryzen Threadripper 7970X, 128GB DDR5 RAM, and an NVIDIA RTX 6000 Ada GPU. For system dynamics and discovery, we deploy a cluster of Raspberry Pi 5 agents. For latency overhead, a Steam Deck OLED acts as a secondary mobile consumer agent.

    \begin{table}[htbp]
        \centering
        \caption{Hardware-accelerated H.264 Encoder Settings for RealSense D435}\label{table:encoderSettings}%
        \begin{tabular}{ll}
            \toprule
            \textbf{Parameter} & \textbf{Value} \\ 
            \midrule
            Resolution & $640 \times 360$ \\
            Framerate & $30$fps \\
            Bitrate & $4$Mbps (CBR) \\
            GOP Size & $30$ frames \\
            Encoder & \verb|h264_nvmpi| \\ 
            \bottomrule
        \end{tabular}
    \end{table}
    
    The edge robots and the Raspberry Pi fleet connect wirelessly to a quad-band WiFi 6E access point, while the base station interfaces with the access point over a 10Gbit wired backbone. Camera streams are encoded using the hardware-accelerated \verb|h264_nvmpi| encoder via \cite{ffmpegImageTransport} on the Jetson (see Table \ref{table:encoderSettings} for details), while decoding on the workstation and Steam Deck OLED relies on a software-based pipeline.

    For reproducibility, all evaluations were conducted in an indoor lab environment. During the tests, the Unitree Go2 remained stationary, capturing a static scene. Specific measurement configurations were established for each evaluation metric:

    \begin{itemize}
        \item \textbf{Bandwidth \& Resource Optimization:} The workstation acted as the consuming agent. Metrics were captured via \verb|psutil| and \verb|iptables|, computing the mean and variance across 120 samples taken at single-second intervals per scaling iteration.
        
        \item \textbf{System Dynamics \& Discovery:} The Raspberry Pi clusters served as the fleet of agents, each executing 10 publisher and 10 subscriber nodes (\verb|demo_nodes_cpp|) to simulate a baseline ROS 2 network. We scale the fleet size progressively, capturing 120 samples per configuration. Prior to each run, all nodes are stopped and the agents sit idle. A central orchestrator broadcasts a simultaneous TCP start signal that immediately executes the ROS 2 launch sequence across the fleet. Discovery time is logged from this start signal until all peers are resolved - evaluated via ROS graph polling in the standard ROS 2 baseline, and via \verb|AgentDiscoverer| discovery events in SFG-ROS. Concurrently, network traffic is quantified using \verb|iptables| to isolate and record peer-specific ingress bytes during the discovery window.

        \item \textbf{Latency Overhead:} The Steam Deck OLED served as the consumer, rendering the camera feed of the edge agent. Glass-to-glass latency was measured as the temporal offset between the activation of a flashing LED positioned in front of the camera and its subsequent detection on the display via a phototransistor. Subscription-to-first-frame latency was measured by blanking an initially illuminated display at the moment of subscription (triggering \verb|AgentDecoder| instantiation) and measuring the duration until the first frame of a continuously lit LED was rendered, as detected by the phototransistor. Both metrics were recorded over 120 samples using an oscilloscope. 
    \end{itemize}

    \subsection{Bandwidth \& Resource Optimization} \label{subsection:bandwidthAndResourceOptimization}

    A primary bottleneck in default ROS 2 implementations is the redundant processing and network transmission when multiple local consumers subscribe to a remote compressed sensor stream. Therefore, we evaluate the computational and network overhead on the consumer device as the number of local subscribers scales from 1 to 10. 

    \begin{figure}[htbp]
        \centering
        \includegraphics{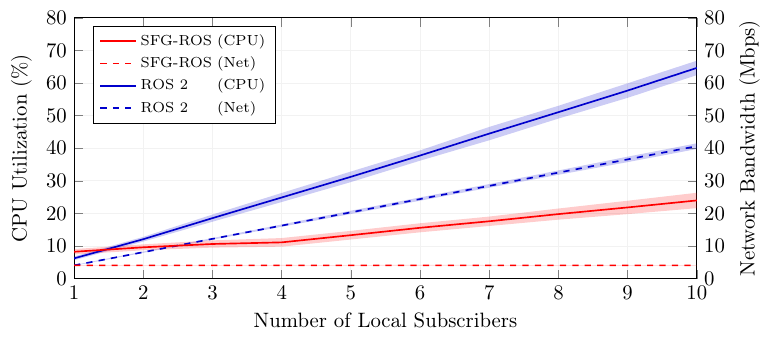}
        \caption{Comparison of CPU utilization and network bandwidth scaling between SFG-ROS and standard ROS 2. CPU utilization is measured such that 100\% represents the full saturation of a single logical CPU core. Shaded regions denote $\pm 1$ standard deviation}\label{figure:resourceOptimization}%
    \end{figure}
    
    As depicted in Figure \ref{figure:resourceOptimization}, standard ROS 2 exhibits an $\mathcal{O}(|\mathcal{S}|)$ linear increase in both CPU utilization and network bandwidth. Specifically, an overhead equivalent to 6.5\% of a single logical CPU core is added for every additional local subscriber due to redundant decompression. Conversely, by deploying the \verb|AgentDecoder|, SFG-ROS completely decouples and bounds the network bandwidth to a constant $\mathcal{O}(1)$ cost (approx. 4Mbps), regardless of subscriber count. While overall CPU utilization remains $\mathcal{O}(|\mathcal{S}|)$ due to local data distribution, the linear growth factor is drastically flattened. As a result, the per-subscriber CPU penalty is reduced to 1.8\% -- a 72.3\% reduction in scaling overhead.
    
    \subsection{System Dynamics \& Discovery} \label{subsection:systemDynamicsAndDiscovery}

    To assess performance in dense deployments, we evaluate both discovery time and network traffic as the number of agents scales from 2 to 6.
    
    \begin{figure}[htbp]
        \centering
        \includegraphics[width=\linewidth]{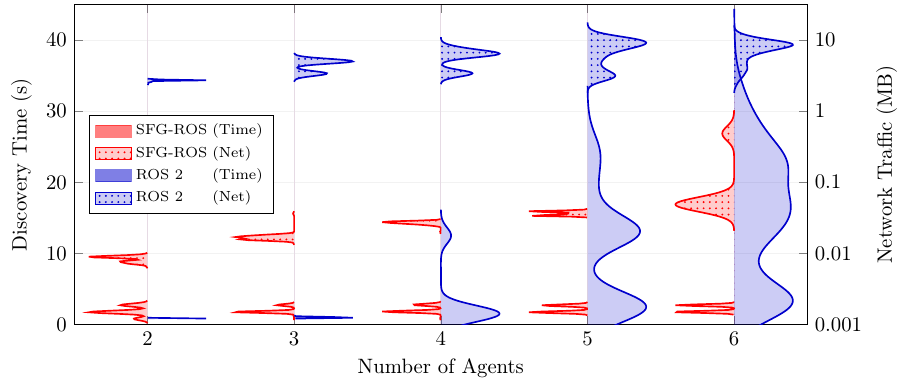} 
        \caption{Fleet discovery metrics across a scaling number of agents. Solid violins map to linear discovery time; dotted violins map to log-scale network traffic. Each distribution is fit to the 120 independent samples taken per configuration}\label{figure:discoveryScaling}%
    \end{figure}

    Figure \ref{figure:discoveryScaling} presents both metrics captured via the testing scenario outlined in Subsection \ref{subsection:experimentalSetup}. Standard ROS 2 relies on multicast UDP discovery, which induces packet storms as node count increases, effectively delaying application-level readiness to up to 40s and consuming up to 13.6MB of network traffic at 6 agents. In contrast, by segregating inter- and intra-agent traffic, SFG-ROS maintains a nearly constant discovery time of 0.8s to 2.9s across scaling iterations and additionally exhibits 1.8 orders of magnitude lower network traffic at the maximum evaluated fleet size.
    
    \subsection{Latency Overhead}
    
    We assess the latency introduced by the framework across two dimensions: the steady-state glass-to-glass delay and the one-time cost of on-demand decoder instantiation.

    \begin{figure}[htbp]
        \centering
        \includegraphics{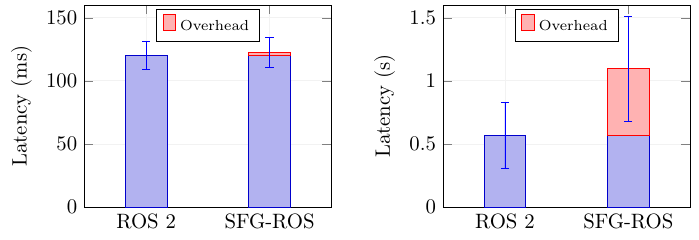}
        \caption{Latency Overhead. Glass-to-glass latency (left) and subscription-to-first-frame latency (right) comparisons between SFG-ROS and standard ROS 2. Error bars denote ±1 standard deviation}\label{figure:latency}%
    \end{figure}

    Figure \ref{figure:latency} (left) illustrates the incurred overhead in glass-to-glass latency measured between the sensor stream from the edge agent to the Steam Deck OLED. Standard ROS 2 transports data directly to the consumer node, yielding a baseline latency of 118.3ms. SFG-ROS introduces a localized relay step (republication via the \verb|AgentDecoder|) and consequently adds a latency overhead of 2.6ms.
    
    Additionally, Figure \ref{figure:latency} (right) illustrates the subscription-to-first-frame latency on the Steam Deck OLED while the edge agent is already discovered. Standard ROS 2 relies solely on native endpoint matching, resulting in a mean latency of 570ms. In contrast, on demand decoder instantiation by SFG-ROS via the \verb|AgentDecoder| increases the mean latency to 1095ms, multiplying initial setup latency by a factor of approximately 1.9.
    
    \section{Conclusion and Future Work} \label{section:ConclusionAndFutureWork}

    We presented the resource-aware, edge-assisted multi-agent software framework SFG-ROS designed to enable robust collaborative perception for heterogeneous robotic fleets. Our topology-enforcing communication architecture leverages a programmatic FQN schema alongside targeted DDS routing to overcome the scalability and resource bottlenecks of default ROS 2 deployments. Coupled with the \verb|AgentDecoder| -- an on-demand, centralized stream decoding pipeline -- this approach significantly reduces redundant computational and network overhead on resource-constrained edge devices. Furthermore, our hardware-agnostic container deployment strategy and specialized spatiotemporal synchronization tools, such as the Vicon Bridge, establish a reproducible and zero-touch foundation from development to field execution. To foster further research and community adoption, the SFG-ROS software package is published under a permissive license and is publicly available via \href{https://iis-esslingen.github.io/sfg-ros}{iis-esslingen.github.io/sfg-ros}.

    While SFG-ROS successfully establishes a robust foundation for multi-agent perception, deploying these systems at greater scale introduces lingering operational challenges. First, while our schema-driven DDS routing effectively isolates local traffic, the underlying DDS protocol still maintains inherent transport and discovery overheads when bridged across volatile, high-density wireless networks. As fleet sizes grow, the underlying middleware must become more inherently resilient to intermittent connectivity and edge-network topologies. Second, while our modular container architecture successfully abstracts heterogeneous hardware, executing agents via mounted \verb|colcon| workspaces remains a development-centric paradigm. Managing raw source code across a fleet lacks the robust versioning and immutability required for reliable, over-the-air production updates. Accordingly, our future work will focus on the following areas:

    \begin{itemize}
        \item \textbf{Middleware Optimization:} Evaluate non DDS-based middlewares (e.g. Eclipse Zenoh) and their native routing capabilities as an alternative to assess potential latency reductions and bandwidth optimizations in dense, multi-agent communication topologies.

        \item \textbf{Automated Deployment Pipelines:} Integrate a CI/CD pipeline for automated \verb|colcon| package building and publishing to a custom Debian repository. This will enable an \verb|apt|-based deployment model within the container's application layer, decoupling source builds from the production runtime while utilizing workspace overlays for development.
    \end{itemize}

    \begin{appendices}

    \end{appendices}

    \section*{Declarations}

    \subsection*{Author Contributions}

    Constantin Blessing conceptualized and developed the SFG-ROS framework, conducted the evaluations for Subsections \ref{subsection:bandwidthAndResourceOptimization} and \ref{subsection:systemDynamicsAndDiscovery}, and wrote the main manuscript. Elias Geiger led the practical validation -- including deployment on the Steam Deck OLED -- and designed and executed the latency evaluation. All authors reviewed the manuscript, provided feedback, and contributed to the refinement of the paper.

    \subsection*{Funding}
    
    Funded by Deutsche Forschungsgemeinschaft (DFG, German Research Foundation) - Projectnumber 528745080 (FIP 68).

    \subsection*{Competing Interests}
    
    The authors have no competing interests to declare that are relevant to the content of this article.

    \subsection*{Data Availability}

    No datasets were generated or analyzed. 

    \subsection*{Code Availability}
    
    Code available via \href{https://iis-esslingen.github.io/sfg-ros}{iis-esslingen.github.io/sfg-ros}.
    

    \bibliography{sn-bibliography}

    \section*{Author Biographies}
    
    \begin{minipage}[t]{0.25\textwidth}
        \vspace{0pt}
        \includegraphics[width=\linewidth]{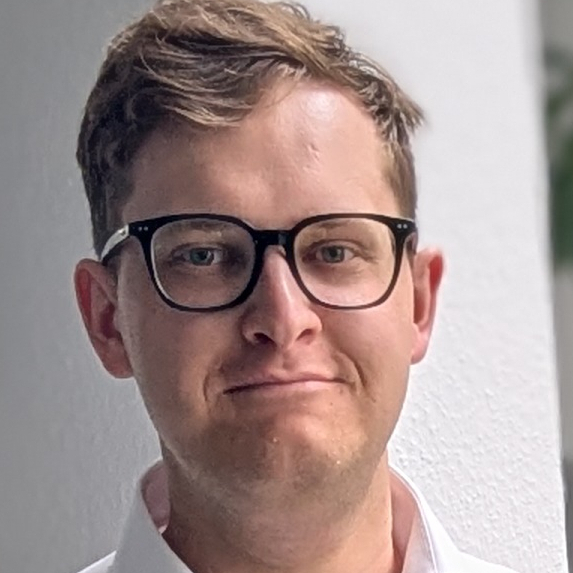}
    \end{minipage}
    \hfill
    \begin{minipage}[t]{0.7\textwidth}
        \textbf{Constantin Blessing} received his M.Sc. degree in applied computer science from the Esslingen University of Applied Sciences, Esslingen am Neckar, Germany, in 2023, where he is currently working toward his Ph.D. degree in computer science supervised by Markus Enzweiler. His research focuses on collaborative task accomplishment and perception by multi-agent systems in smart factory environments.
    \end{minipage}
    \vspace{2em}

    \noindent
    \begin{minipage}[t]{0.25\textwidth}
        \vspace{0pt}
        \includegraphics[width=\linewidth]{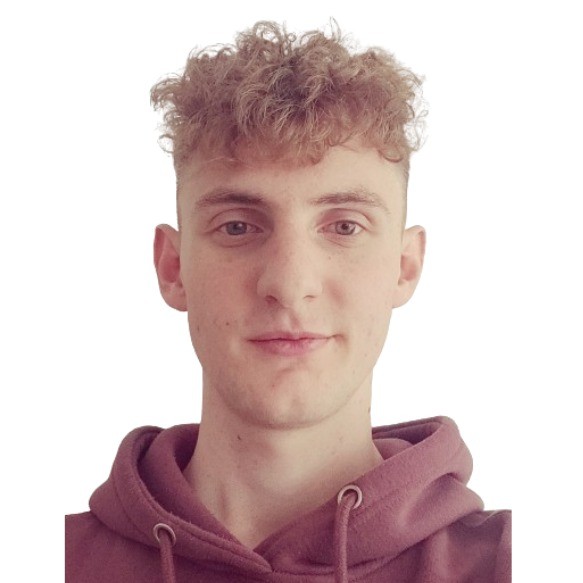}
    \end{minipage}
    \hfill
    \begin{minipage}[t]{0.7\textwidth}
        \textbf{Elias Geiger} Elias Geiger is a bachelor’s student in Technical Computer Science at Esslingen University. His academic interests focus on robotics, embedded systems, and the application of computer science methods to intelligent systems.
    \end{minipage}
    \vspace{2em}
    
    \noindent
    \begin{minipage}[t]{0.25\textwidth}
        \vspace{0pt}
        \includegraphics[width=\linewidth]{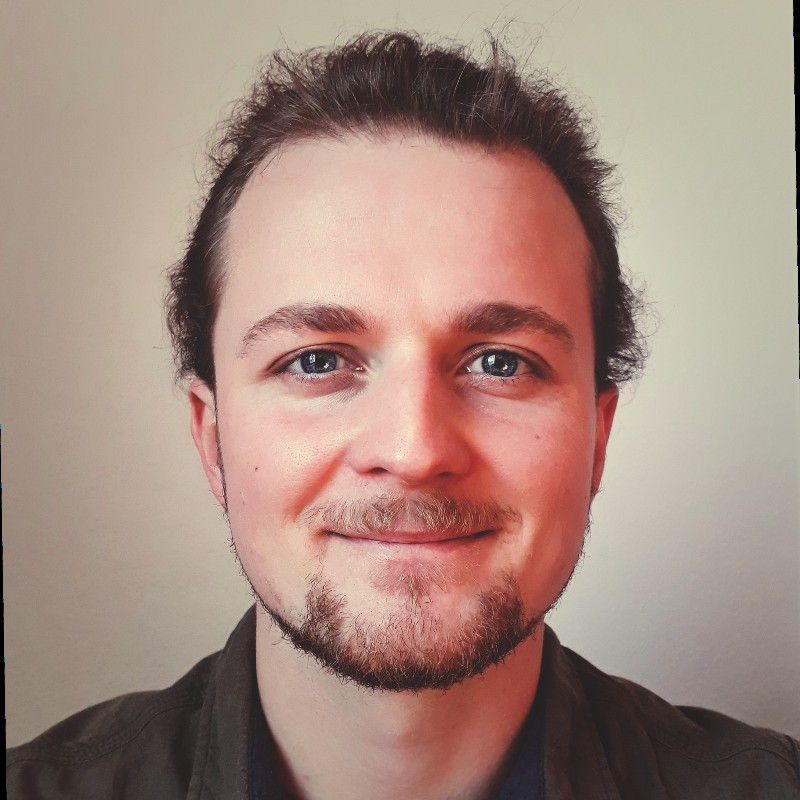}
    \end{minipage}
    \hfill
    \begin{minipage}[t]{0.7\textwidth}
        \textbf{Jakob Häringer} received his M.Sc. degree in applied computer science from the Esslingen University of Applied Sciences, Esslingen am Neckar, Germany, in 2023, where he is currently working toward his Ph.D. degree in computer science supervised by Markus Enzweiler. His research focuses on robotic manipulation in smart factory environments, specifically the use of Vision-Language-Action models for the handling of uncommon objects.
    \end{minipage}
    \vspace{2em}
    
    \noindent
    \begin{minipage}[t]{0.25\textwidth}
        \vspace{0pt}
        \includegraphics[width=\linewidth]{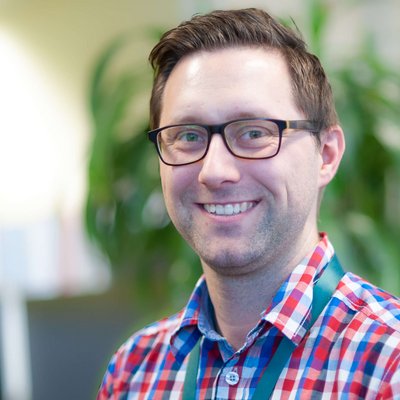}
    \end{minipage}
    \hfill
    \begin{minipage}[t]{0.7\textwidth}
        \vspace{0pt}
        \textbf{Dennis Grewe} received his Ph.D. degree in computer science from the University of Koblenz, Koblenz, Germany, in 2021. From 2019 to 2023, he was a Research Engineer with Robert Bosch GmbH. Since September 2024, he has been a Full Professor of Distributed Systems and Software Testing with the Esslingen University of Applied Sciences, Esslingen am Neckar, Germany. His research interests include network protocols and architectures to support computing on network infrastructures, including principles from software-defined, agentic, and data-oriented networking for distributed systems.
    \end{minipage}
    \vspace{2em}

    \noindent
    \begin{minipage}[t]{0.25\textwidth}
        \vspace{0pt}
        \includegraphics[width=\linewidth]{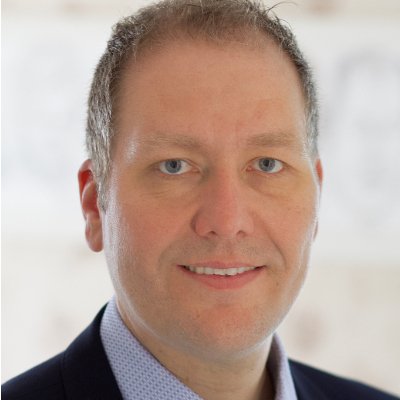}
    \end{minipage}
    \hfill
    \begin{minipage}[t]{0.7\textwidth}
        \textbf{Markus Enzweiler} received his M.Sc. degree in computer science from the University of Ulm, Ulm, Germany, in 2005, and his Ph.D. degree in computer science from the University of Heidelberg, Heidelberg, Germany, in 2011. In 2002 and 2003, he was a Visiting Student Researcher with the Centre for Vision Research, York University, Toronto, ON, Canada. From 2010 to 2020, he has been with Mercedes-Benz Research and Development, Stuttgart, Germany, focusing on camera- and LiDAR-based scene understanding for self-driving cars. Since 2021, he has been a Full Professor of Computer Science at Esslingen University of Applied Sciences, Esslingen am Neckar, Germany, where he founded and heads the Institute for Intelligent Systems.
    \end{minipage}
\end{document}